\documentclass[letterpaper, 10 pt, conference, hidelinks]{ieeeconf}  
\IEEEoverridecommandlockouts
\usepackage{hyperref}
\usepackage{graphics} 
\let\labelindent\relax
\usepackage{epsfig} 
\usepackage{xcolor}
\hypersetup{
    colorlinks,
    linkcolor={black!50!black},
    citecolor={blue!50!black},
    urlcolor={blue!80!black}
}
\usepackage{amsmath} 
\usepackage{mathtools}
\usepackage{amssymb}  
\usepackage{enumitem}
\usepackage{comment}
\usepackage{tabularx}
\usepackage{algorithm}
\usepackage{algpseudocode}
\usepackage{multirow}
\usepackage{booktabs} 
\usepackage{graphicx} 
\usepackage{enumitem} 
\usepackage{hyperref} 
\usepackage{array}
\usepackage{float}
\usepackage{subcaption}
\usepackage{etoolbox}
\usepackage{cite}

\definecolor{burntorange}{rgb}{0.8, 0.33, 0.0}

\newcolumntype{P}[1]{>{\centering\arraybackslash}p{#1}}

\begin{document}
\newcommand{\Navo}[1]{\textcolor{blue}{\textit{NK: #1}}}
\title{\LARGE \bf Learned Slip-Detection-Severity Framework using Tactile Deformation Field Feedback for Robotic Manipulation}
\newcommand{\etal}{\textit{et al. }}

\author{\authorblockN{Neel Jawale\authorrefmark{1}\authorrefmark{2},
Navneet Kaur\authorrefmark{1}\authorrefmark{2}, Amy Santoso\authorrefmark{2}, Xiaohai Hu\authorrefmark{2}, Xu Chen\authorrefmark{2} \authorrefmark{3}%
\thanks{This work is supported in part by NSF Award \#2141293. \authorrefmark{2}:Authors are with the MACS lab, Department of Mechanical Engineering, University of Washington,
       Seattle, WA 98195, USA. {\tt\small\{neelj42, navneet, esantoso, huxh, chx\}@uw.edu}. \authorrefmark{1}: equal contribution \authorrefmark{3}: corresponding author }%
}}%
\maketitle
\pagestyle{empty}
\begin{abstract}
Safely handling objects and avoiding slippage are fundamental challenges in robotic manipulation, yet traditional techniques often oversimplify the issue by treating slippage as a binary occurrence. Our research presents a framework that both identifies slip incidents and measures their severity. We introduce a set of features based on detailed vector field analysis of tactile deformation data captured by the GelSight Mini sensor. Two distinct machine learning models use these features: one focuses on slip detection, and the other evaluates the slip's severity, which is the slipping velocity of the object against the sensor surface. Our slip detection model achieves an average accuracy of 92\%, and the slip severity estimation model exhibits a mean absolute error (MAE) of 0.6 cm/s for unseen objects. To demonstrate the synergistic approach of this framework, we employ both the models in a tactile feedback-guided vertical sliding task. Leveraging the high accuracy of slip detection, we utilize it as the foundational and corrective model and integrate the slip severity estimation into the feedback control loop to address slips without overcompensating. Videos and demonstrations are available at: \url{https://sites.google.com/uw.edu/lsds}

\end{abstract}

\section{INTRODUCTION}
Tactile sensing plays a pivotal role in robotic manipulation, offering a rich source of information for understanding and interacting with the environment \cite{lee2019making}. Manipulation tasks such as delicate handling of objects \cite{soft_tactile} and secure grasping \cite{single_grasp} under dynamic conditions rely on effective slip detection \cite{friction_tactile}. Traditional approaches to slip detection have predominantly relied on binary indicators of slip occurrence \cite{veiga2, proactive, Li2018SlipDW, james1, hu2023learning}, leveraging tactile data to discern stable grips from unstable ones. However, this binary treatment overlooks the nuanced spectrum of slip dynamics, potentially leading to inadequate control strategies. Furthermore, reliance solely on visual feedback for slip detection and manipulation tasks presents inherent limitations, such as occlusions, varying lighting conditions, and the need for external viewpoints that may not always be feasible in confined environments \cite{ebert2018visual}.

\begin{figure*}[bt]
    \centering
    \includegraphics[width=0.75\linewidth, height=130pt]{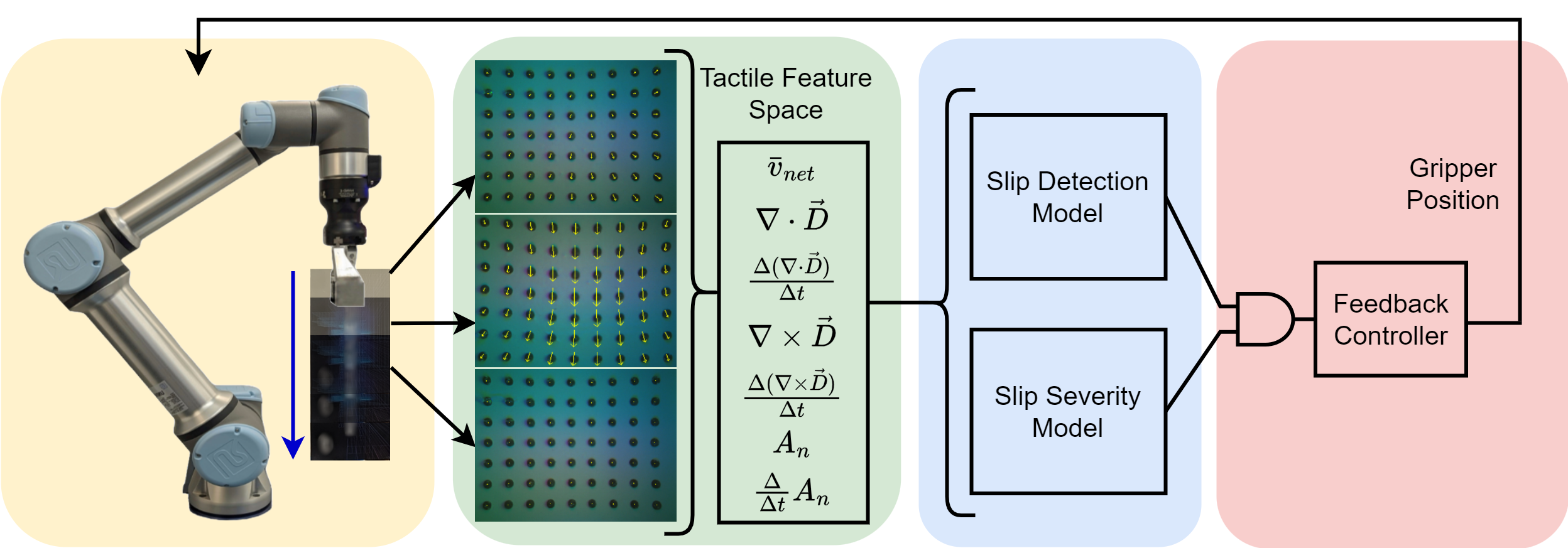}
    \caption{Summary of the Slip-Detection-Severity Framework: A robot executes an object handling task, during which tactile features from the GelSight Mini sensor are extracted in real time. These features simultaneously feed into the Slip Detection and Slip Severity models. Upon detecting slip, the feedback controller actively adjusts the gripper to mitigate slip severity.}
    \label{fig:mainfig}
    \vspace{-10pt}
\end{figure*}

To overcome the identified limitations, this research seeks to answer the primary question of how slip detection methods can be improved to effectively manage the complexities of slip dynamics. Furthermore, it explores how can slip be quantified to aid in the development of feedback control algorithms for more precise slip handling and mitigation.

Our contribution centers on the direct detection of slip occurrences and the simultaneous estimation of slip severity, utilizing real-time tactile sensor data.
We introduce a framework for learned slip detection and severity assessment, derived from the tactile features of the deformation vector field extracted from GelSight sensors. By identifying essential tactile features that capture the non-linear surface dynamics associated with slip events, we employ machine learning to directly map these features for slip detection and severity assessment.  The efficacy of our framework is demonstrated through its integration into a feedback gripper controller as shown in Figure \ref{fig:mainfig}, showcasing how the slip-detection-severity feedback enables precise gripper positioning without the need for explicit force control or prior knowledge of the object's size, geometry, or texture.

\textit{Background and Related Work:} Research in tactile sensing was initiated in the 1970s with the introduction of piezoelectric elements as strain sensors \cite{KINOSHITA1975}, and has expanded to include a diverse array of sensor technologies. These technologies are capable of detecting various object properties such as mass, geometry, texture, slip, and hardness, utilizing piezoelectric, capacitive array, optical, and magnetic sensors \cite{siciliano}. Among these, optical marker-based tactile sensors, including TacTip \cite{ward}, TouchRoller \cite{touchroller}, and GelSight \cite{GelSight}\cite{GelSight2}, represent significant advancements in sensing high-resolution surface features and texture. TacTip sensors employ a camera to monitor the movement of white pins within a membrane upon object contact, mirroring a biomimetic design. TouchRoller, designed as a rolling sensor, acquires geometric information by traversing an object's surface. This study focuses on the GelSight sensor, which uses a reflective gel-coated elastomer and LED illumination to track marker displacement, enabling accurate measurements of contact deformation using the deformation vector field.

Foundational work has been established by demonstrating the use of tactile data for slip detection \cite{howeslip}. Optical methods utilize the eccentricity of the contact surface to measure object deformation \cite{ikeda}.
The integration of machine learning has expanded slip detection capabilities, with Support Vector Machines (SVMs) utilized alongside TacTip sensor data \cite{james_slipdetection}. Neural networks have been applied for slip classification \cite{agriomallos}, and GelSight sensor data have been incorporated to enhance object shape detection \cite{Li2018SlipDW}. Furthermore, entropy-based methods utilize learning algorithms and shear marker displacement to predict slip likelihood \cite{hu2023learning, wenzhen}. In this study, we concentrate on extracting tactile features through vector field analysis of the deformation field and employing learned models to detect slip.

Research on predicting the relative velocity between grippers and manipulated objects through tactile sensing alone remains limited due to the complexity involved. A method for calculating sliding velocity utilizes a nonlinear observer used with the SUNTouch tactile sensor \cite{app13020921}. However, this technique encounters difficulties with objects that have small curvature radii or high deformability. Another approach, employing linear regression and capacitive-based nib-structure tactile sensors for slip speed prediction \cite{gloumakov2022learning}, presents a promising direction despite limitations in the feature space and low correlation with the regression variable. Recent efforts have leveraged CNNs alongside BioTac sensor measurements to assess the slipping speed of objects \cite{chen2021}. This approach is particularly effective with rigid objects but faces challenges in generalizing to deformable ones. This study aims to establish a direct correlation between slip velocity and tactile features extracted from vector field analysis, using learned models. We demonstrate that extracting features from the optical tactile sensors enables the development of models that effectively generalize across different object types.

\section{PROBLEM SETUP}
In the context of this research, ``slip" refers to the relative motion between a grasped object and the tactile sensor surface. Slip can occur due to various factors such as inadequate friction between the gripper and the object, external forces applied to the object, or surface irregularities.
We concentrate on the following three challenges:
\begin{enumerate}[leftmargin=*]
    \item \textbf{Detection:} How can we accurately identify instances of slip occurrence just using features extracted from the tactile sensor?
    \item \textbf{Estimation:} How can we gauge the relative velocity between the object and the sensor using the same features?
    \item \textbf{Control:} How can we integrate the detection and estimation pipelines to efficiently regulate gripper position during manipulative tasks, preventing object slippage?
\end{enumerate}

Our experimental setup incorporates a combination of robotics and tactile sensing technologies as follows:

\begin{enumerate}[leftmargin=*]
    \item An UR5e robotic arm with custom-designed, position-controlled end-effector fingers mounted on a Robotiq Hand-E flange, specifically designed for GelSight Mini sensor integration.
    
    \item A GelSight Mini sensor, chosen for its detailed surface data capture. We depend on tactile feedback from only one of the sensors mounted on the fingers, as it suffices for the research scope and offers computational efficiency.
    
    \item A spatial data acquisition platform utilizing an Intel RealSense D435i camera to gather 3D pose information for collecting ground truth data. 
    
    \item A computational core consisting of a dedicated PC running Ubuntu 20.04 LTS and ROS Noetic, equipped with an Intel i7 CPU and a NVIDIA GeForce RTX 3050 GPU.
\end{enumerate}

\section{METHODOLOGY}
\subsection{Tactile Feature Selection and Rationalization}
\begin{figure}[tb]
    \centering
    \includegraphics[width=0.85\linewidth]{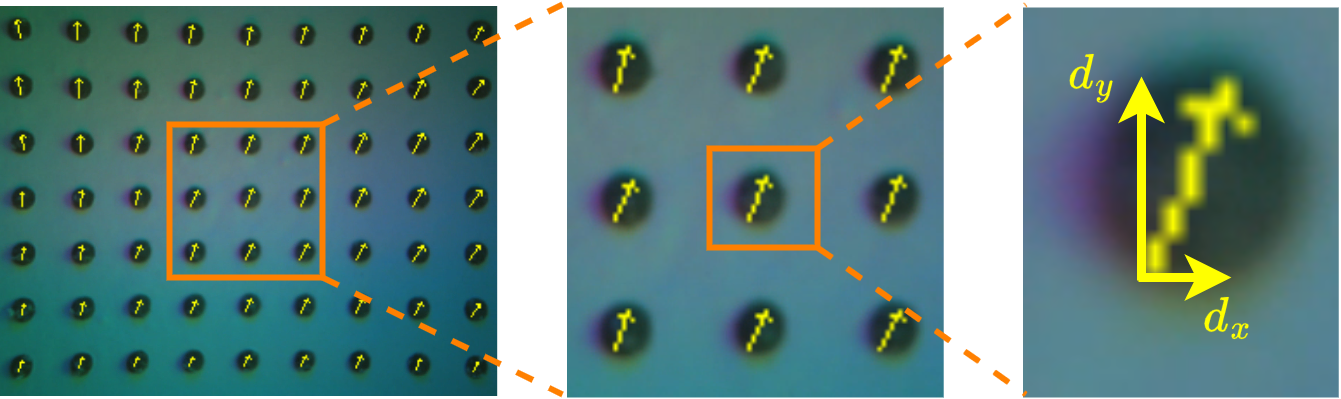}
    \caption{The figure depicts the deformation vector fields through the displacement of markers, highlighting the displacement vector components of a specific marker.}
    \label{fig:tactile_features}
    \vspace{-12pt}
\end{figure}

The GelSight Mini sensor is equipped with a built-in camera that records the deformation of an elastomeric gel surface, capturing it as a sequence of frames. This gel surface is engraved with 63 black markers, enabling precise pixel tracking. By employing optical flow analysis \cite{roboticAssembly}, we measure the deformation field by estimating the movement of these markers across video frames relative to a reference frame depicting the gel's undisturbed state. This methodology provides a detailed measure of surface interactions and deformations. The displacement markers, illustrating the deformation field, are as shown in Figure \ref{fig:tactile_features}. 
\subsubsection{Kinematics of Tactile Deformation Field}
The velocity of the tactile deformation field is essential for detecting rapid contact changes, which occur due to the movement of the gel on the sensor's surface. Velocity is represented as the rate of change in gel displacement, and its components for marker \(i\) in the \(x\) and \(y\) directions are defined as:
\begin{equation}
v_{x_i} = \frac{\Delta d_{x_i}}{\Delta t}, \quad v_{y_i} = \frac{\Delta d_{y_i}}{\Delta t}
\end{equation}
where \( v_{x_i} \) and \( v_{y_i} \) represent the respective velocity components, \(d_{x_i}\) and \(d_{y_i}\) represent the displacement in $x$ and $y$ direction respectively, and \(\Delta t (=0.04 \text{s}) \) is the sampling time of the GelSight sensor.
For the \(N (=63)\) markers engraved on the GelSight mini sensor, the mean velocities are given by:
\begin{equation}
\bar{v}_x = \frac{1}{N} \sum_{i=1}^{N} v_{x_i}, \quad \bar{v}_y = \frac{1}{N} \sum_{i=1}^{N} v_{y_i}
\end{equation}

To effectively capture the overall motion dynamics, we employ the \(L_2\) norm of the velocity components \(v_x\) and \(v_y\) as a key feature in our analysis:
\begin{equation}
\bar{v}_{net} = \sqrt{\bar{v}_x^2 + \bar{v}_y^2}
\end{equation}

\subsubsection{Vector Analysis of Tactile Deformation Field}
\begin{figure}[tb]
\vspace{8pt}
    \includegraphics[width=\linewidth]{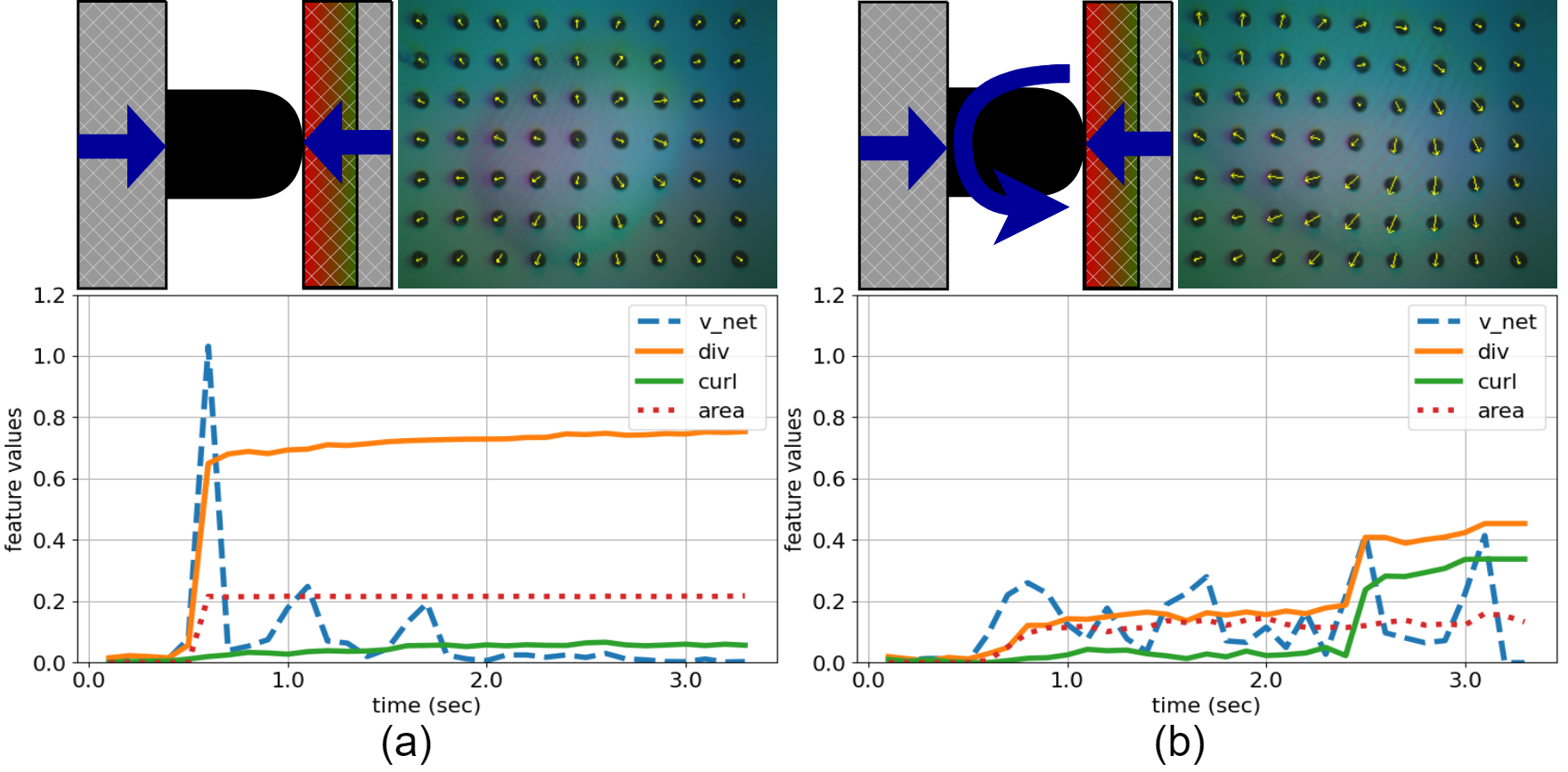}
    \caption{Two tactile interaction scenarios, each with corresponding sensor frames and graphs (sensor mounted on a single finger), are presented: (a) Static grasping, where there is no relative motion between the object and the sensor, divergence increases noticeably. (b) Object rotation, introduces torsional stress, leading to simultaneous increases in both divergence and curl, as well as inhomogeneity in instantaneous area of contact, as observed in the graph.}
    \label{fig:div_curl}
    \vspace{-10pt}
\end{figure}

The deformation vector field provides crucial flow information regarding contact dynamics and forces \cite{roboticAssembly, zhang2019effective}. We propose to employ vector flow analysis to derive additional features, divergence and curl, from discrete vector fields extracted from the GelSight sensor. Divergence is the net rate of expansion or contraction of the deformation field, typically associated with normal force application, while curl reflects the rotational forces at play, as visualized in Figure \ref{fig:div_curl}. Notably, rotational forces not only elevate curl but can also cause divergence to vary, illustrating the interplay between rotational forces and fluctuating normal forces on the sensor's surface. For the discrete deformation vector field extracted from the tactile sensor output, divergence and curl are expressed as:
\begin{equation}
\begin{aligned}
\nabla \cdot \vec{D} &= \sum_{i=1}^{N} \left( \frac{d_{x_{i+1}} - d_{x_{i-1}}}{2\Delta x} + \frac{d_{y_{i+1}} - d_{y_{i-1}}}{2\Delta y} \right)\\
\nabla \times \vec{D} &= \sum_{i=1}^{N} \left( \frac{d_{y_{i+1}} - d_{y_{i-1}}}{2\Delta x} - \frac{d_{x_{i+1}} - d_{x_{i-1}}}{2\Delta y} \right)
\end{aligned}
\end{equation}
where \(\Delta x\) and \(\Delta y\) represent spatial resolution between markers.
Observing the rate of change of these vector quantities enhances our ability to detect and respond to dynamic tactile events, as they reflect changes in the distribution and directionality of forces at the contact surface. These changes can be quantitatively expressed as:
\begin{equation}
\begin{aligned}
\frac{\Delta(\nabla \cdot \vec{D})}{\Delta t} &= \frac{(\nabla \cdot \vec{D})_{t} - (\nabla \cdot \vec{D})_{t-\Delta t}}{\Delta t}\\
\frac{\Delta(\nabla \times \vec{D})}{\Delta t} &= \frac{(\nabla \times \vec{D})_{t} - (\nabla \times \vec{D})_{t-\Delta t}}{\Delta t}
\end{aligned}
\end{equation}

The rate of change of divergence captures how quickly the net deformation field evolves to the normal force being applied on the sensor, signaling a potential loss of grip due to insufficient grasp. Similarly, a sudden increase in the rate of change of curl captures the tangential movements and rotational shifts between the sensor and the object.

\subsubsection{Normalized Contact Area}
The assessment of the slip phenomenon benefits from analyzing the contact area between an object and the tactile sensor and observing how this area changes over time. A rapid change in the contact area indicates disturbance at the sensor surface. The greater the rate of change in the area on the sensor surface, the higher the severity of the interaction. The left graph in Figure \ref{fig:div_curl} illustrates a stable contact area under the normal force indicating a stable grip, while the right graph shows significant fluctuations under combined torsional and normal forces, indicating more complex interactions and slip occurrences. Our study examines this by employing the 3D-Recon technique \cite{GelSight}, which reconstructs the tactile sensor's surface depth using a photometric stereo algorithm. We determine the contact area by counting the pixels where the depth exceeds a carefully selected threshold of 1, which accounts for the gel's micro-variations and ensures precision in our measurements. We define the normalized contact area, \(A_n\), as the proportion of pixels surpassing this threshold relative to the total number of pixels, as shown in the following equation:
\begin{equation}
A_n = \frac{\sum\limits_{i=1}^{m} \sum\limits_{j=1}^{n} I(z_{ij} > z_t)}{m \times n}\\
\end{equation}
Here, \(I\) is an indicator function that activates when the depth \(z_{ij}\) at any pixel exceeds the threshold \(z_t\), with \(m(=320)\) and \(n(=240)\) representing the sensor image dimensions. The time derivative of \(A_n\) is defined as:

\begin{equation}
\frac{\Delta A_n}{\Delta t} = \frac{A_n(t) - A_n(t-\Delta t)}{\Delta t}
\end{equation}

\subsubsection{Baseline features}
As mentioned in \cite{wenzhen} and \cite{hu2023learning}, entropy and its rate of change serve as reliable indicators of slip. The entropy of the deformation field, as described in these works, is defined as the statistical measure of randomness in the distribution of marker displacements. Mathematically, for a discrete deformation field, such as that extracted from the GelSight sensor, entropy can be represented using Shannon entropy:
\begin{equation}
H(X) = -\sum_{x \in X} p(x) \log p(x)
\end{equation}
where \(X\) denotes the histogram representation of marker displacements \cite{funkhouser2012}. Entropy (\(H\)) and its rate of change (\(\frac{\Delta H}{\Delta t}\)) serve as baseline features against which we compare the proposed vector field features in tasks related to slip detection and severity estimation.

\subsection{Slip Detection Model}
Utilizing quantifiable and physical features extracted from the tactile sensor facilitates the application of statistical theory-based algorithms, such as classifiers, in slip detection endeavors \cite{ml_tactile}. Commonly employed algorithms in this domain include Support Vector Machines (SVM), K-Nearest Neighbors (KNN), Decision Trees (DT), Random Forest (RF), and Linear Discriminant Analysis (LDA). Research has shown that ensemble tree algorithms, particularly when bagged or boosted, are effective in slip detection because of their rapid processing and excellent performance with non-linear data \cite{Levins2020ATS}. 
Hence, our study will concentrate on using Random Forest (RF) \cite{breiman2001random} and Gradient Boosting (GB) \cite{friedman2001greedy} as our slip detection models, evaluating their effectiveness against baseline and newly proposed tactile features. We utilize the Scikit-learn library to implement these classifiers for analysis and inferencing.

\subsection{Slip Severity Estimation Model}
Once a slip detection model is learned and evaluated, we focus on quantifying slip severity. The quantification is based on understanding the physics behind the tactile deformation field and the velocities of objects slipping against sensor surfaces. We observed that a higher velocity of slippage often correlates with an increased difficulty in regaining control, suggesting a higher severity of slip. Our approach seeks to establish a correlation between tactile sensory features and object velocity, enabling the prediction of slip severity using data from tactile sensors alone. This method could be particularly useful in environments where external vision-based systems are not feasible. We employ two deep learning-based models using PyTorch: a Long Short-Term Memory (LSTM) network \cite{hochreiter1997} and a Multilayer Perceptron (MLP) \cite{haykin1994neural}. These models are chosen to explore both the sequential nature of the data and the relationships among static features at discrete time points, respectively. We then compare the performance of these models using baseline and proposed tactile features.

\subsection{Slip-Detection-Severity Feedback Gripper Control}
\begin{algorithm}
\caption{PD Control for Slip Mitigation using HandE gripper}
\begin{algorithmic}[1]
\Statex \textbf{Inputs:} $S_{\text{slip}}$ : Slip detection signal 
\Statex $S_{\text{severity}}$ : Slip severity 
\Statex $p_{\text{current}}$ : Current gripper position 
\Statex $S_{\text{target}} = 0$ : Desired slip severity
\Statex $K_p$ : Proportional gain 
\Statex $K_d$ : Derivative gain
\Statex \textbf{Output:} $p_{\text{new}}$ : New gripper position
\Statex \textbf{Initialize:} $e_{\text{previous}} \gets 0$ 
\While{Gripper is operational}
    \If{$S_{\text{slip}}$ is detected}
        \State $e \gets S_{\text{severity}} - S_{\text{target}}$
        \State $\dot{e} \gets e - e_{\text{previous}}$
        \State $p_{\text{adjustment}} \gets K_p \cdot e + K_d \cdot \dot{e}$
        \State $p_{\text{new}} \gets \max(\min(p_{\text{current}} - p_{\text{adjustment}}, 225), 0)$
        \State Update gripper position to $p_{\text{new}}$
        \State $e_{\text{previous}} \gets e$
    \Else
        \State Maintain current gripper position
    \EndIf
\EndWhile
\end{algorithmic}
\end{algorithm}

With the development of the slip detection and slip severity models, we further implement a Proportional-Derivative (PD) controller on the Robotiq Hand-E gripper, which solely manipulates the gripper's position to adjust gripping force. The controller enhances grip by modulating the distance between the gripper fingers: the Robotiq Hand-E is fully open at a position value \(p\) of 0 and completely closed at 225. This control strategy, alongside the slip severity estimator, is activated upon slip detection to mitigate slip severity and is otherwise inactive to conserve computational resources and avoid unnecessary adjustments. Once slip velocity is ascertained by the estimation model, the gripper controller intervenes to enhance grip, thereby aiming to reduce the slip severity to zero. This approach demonstrates that effective control can be achieved using a straightforward position-controlled gripper within the proposed framework.

\section{EXPERIMENTS AND RESULTS}
To evaluate the effectiveness of the slip detection and slip severity models discussed in the previous section, both individually and in combination as a learned slip-detection-severity framework, we conducted three experiments. The first experiment involved deploying the developed slip detection models and assessing their performance through comparative analysis. The second experiment was dedicated to implementing and evaluating the slip severity models, utilizing the same flow features extracted from the tactile deformation field for slip detection. The third experiment explored a synergistic approach to slip detection and severity assessment by integrating a feedback Proportional Derivative (PD) gripper controller for vertical object sliding downstream task. We used the same features for both Slip Detection and Slip Severity. These features are categorized throughout the text as follows:

\begin{description}
    \item[Baseline:] \(H\) and \(\frac{\Delta H}{\Delta t}\)
    
    \item[Proposed:] \(\bar{v}_{net}\), \(\nabla \cdot \vec{D}\), \(\nabla \times\vec{D}\), \(\frac{\Delta(\nabla \cdot \vec{D})}{\Delta t}\), \(\frac{\Delta(\nabla \times \vec{D})}{\Delta t}\), \(A_n\), \(\frac{\Delta A_n}{\Delta t}\)
    
    \item[Combined:] Baseline + Proposed
\end{description}

\subsection{Slip Detection}
\subsubsection{Experimental Setup and Data Collection}
\begin{figure}[tb]
\vspace{3pt}
    \centering
    \includegraphics[width=0.8\linewidth]{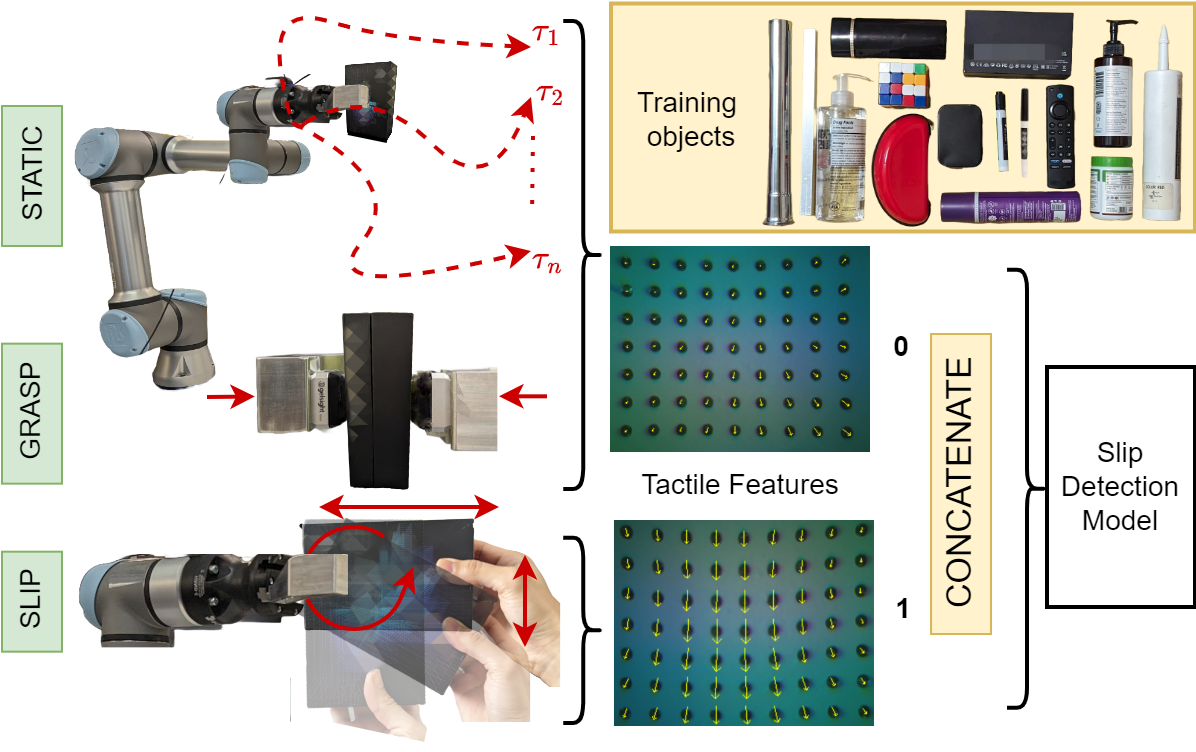}
    \caption{Schematic representation of the data collection pipeline for the Slip Detection experiment. The STATIC and GRASP scenarios detail the methodology for acquiring `no-slip' data labeled 0, while the SLIP scenario illustrates the process for gathering data indicative of slip labeled 1. Both datasets are combined during the training phase for input into the slip detection model.}
    \label{fig:ML_exp}
    \vspace{-10pt}
\end{figure}

Data collection for slip detection was organized into three scenarios: static, grasp, and slip, elaborated in Figure \ref{fig:ML_exp}. In the static scenarios, the robot was programmed using the MoveIt framework to follow 10 unique trajectories, each with 5 increasingly tighter grasps. This method aimed to document varying vector field intensities for each of the 15 objects selected, striving for precise spatial poses without inducing slip. The recording duration was standardized across scenarios and objects to ensure consistency, with each object undergoing identical trajectories. For the grasp scenario, objects were sequentially grasped, moving from an initial gripper position \(p\), representing minimal contact, to a final position \(p+10\), pausing for 5 seconds before each position increment. Both static and grasp scenarios were marked with a label of `0', indicating no slip occurred. In the slip scenarios, slippage was deliberately caused while objects were in contact with the GelSight sensors. This setup enabled the collection of data on various slip intensities over identical timeframes to those in the no-slip experiments, with this data being assigned a label of `1'. Data for each object was collected once per category, with each recording lasting approximately one minute. The dataset was annotated through manual labeling by human observers, to effectively differentiate between the distinctions of static and slipping states. Following the concatenation process, our training dataset expanded to approximately 66,000 data points. In this dataset, all columns except for the last serve as feature values, the final column represents the annotation, and each row is one data point.

\subsubsection{Results}
\paragraph{Metrics used for comparison}
In assessing the classification models' effectiveness in this study, a diverse set of metrics was utilized to thoroughly evaluate their performance. These metrics, summarized in Table \ref{tab:performance_metrics}, aimed to capture various aspects of the model's efficacy. Here, True Positives (TP) are instances correctly predicted as positive, True Negatives (TN) are instances correctly predicted as negative, False Positives (FP) are instances incorrectly predicted as positive, and False Negatives (FN) are instances incorrectly predicted as negative.
\begin{table}[h]
\vspace{5pt}
\centering
\caption{\label{tab:performance_metrics}Classification metrics summary}
\setlength\extrarowheight{5pt}
\begin{tabularx}{\columnwidth}{|l|l|X|}
\hline
\textbf{Metric} & \textbf{Definition}                                  & \textbf{Formula}                               \\ \hline
Accuracy (A)            & Overall correctness                                  & $\frac{TP + TN}{TP + TN + FP + FN}$            \\
Precision (P)          & Positives' accuracy                                  & $\frac{TP}{TP + FP}$                           \\
Recall (R)           & Positive instance identification                     & $\frac{TP}{TP + FN}$                           \\
F1 Score (F)            & Precision \& recall balance                          & $2 \times \frac{\text{Precision} \times \text{Recall}}{\text{Precision} + \text{Recall}}$ \\
\hline
\end{tabularx}
\end{table}

\paragraph{Cross-validation for generalization}
We conducted hyperparameter tuning for RF and GB, setting specific values for parameters such as \textit{max\_depth}, \textit{max\_features}, \textit{min\_samples\_leaf}, \textit{min\_samples\_split}, and \textit{n\_estimators}. For RF, we chose values of 20, 3, 5, 10, and 40, while for GB, we selected the first three parameters to be 9, 450, and 115 (with the rest set to default). These models were then trained on all feature categories - baseline, proposed, and combined. 
We evaluated the models using Stratified 5-fold cross-validation to ensure fair assessment across both classes. Results presented in Table \ref{tab:stratkfoldresults} indicate that with the proposed features, RF and GB achieved a mean accuracy of 99.11\% and 99.22\%, respectively. Combining these with baseline features resulted in a slight increase in accuracy - 0.26\% for RF and 0.35\% for GB, suggesting that the proposed features effectively capture the relationship with slip occurrences.

\begin{table}[tb]
\centering
\caption{Stratified 5-fold cross-validation results}
\setlength\extrarowheight{5pt}
\label{tab:stratkfoldresults}
 \begin{tabularx}{\columnwidth}{| c | c | >{\centering\arraybackslash}X | >{\centering\arraybackslash}X | >{\centering\arraybackslash}X | >{\centering\arraybackslash}X |}
  \hline
  \multicolumn{2}{|c|}{\textbf{Metrics}} & A & P & R & F \\
  \hline
  \multirow{3}{*}{RF} & Baseline & 93.94 & 94.83 & 92.31 & 93.55 \\ \cline{2-6}
  & Proposed & 99.11& 98.76 &  99.11 & 98.93 \\ \cline{2-6}
  & \textbf{Combined} & \textbf{99.37 }   & \textbf{99.1} &  \textbf{99.38 }    & \textbf{99.24} \\ 
  \hline
  \multirow{3}{*}{GB} & Baseline & 93.97  & 94.75  & 92.46 & 93.59 \\ \cline{2-6}
  & Proposed & 99.22    & 98.97 &  99.17     & 99.07 \\ \cline{2-6}
  & \textbf{Combined} & \textbf{99.57}    & \textbf{99.49} &  \textbf{99.47} & \textbf{99.48} \\
  \hline
 \end{tabularx}
 \vspace{-8pt}
\end{table}

\begin{table}[b]
\vspace{-8pt}
\centering
\setlength\extrarowheight{3pt}
\caption{\label{table:ML_test}Results of trained classifiers on unseen objects}
\begin{tabularx}{\columnwidth}{|ll|l|>{\centering\arraybackslash}X|>{\centering\arraybackslash}X|>{\centering\arraybackslash}X|>{\centering\arraybackslash}X|}
\hline
\multicolumn{3}{|c|}{{\textbf{Metrics}}} & A & P & R & F\\
\hline
\multirow{2}{*}{\includegraphics[width=0.65cm]{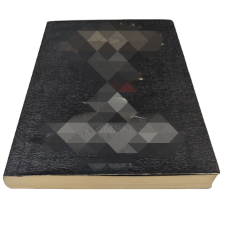}} & \multirow{2}{*}{Book}
& \textbf{RF} & \textbf{91.76} & \textbf{84.73} & \textbf{98.27} & \textbf{91.0} \\
\cline{3-7}
&& GB & 83.71 & 83.74 & 98.96 & 72.58 \\
\hline
\multirow{2}{*}{\includegraphics[width=0.65cm]{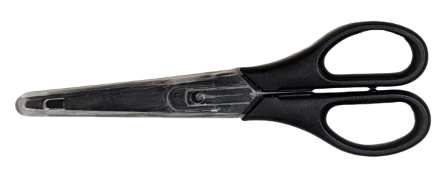}} & \multirow{2}{*}{Scissors}
& RF & 99.51 & 99.28 & 99.54 & 99.41 \\
\cline{3-7}
&& \textbf{GB} & \textbf{99.57} & \textbf{99.44} & \textbf{99.54} & \textbf{99.49} \\
\hline
\multirow{2}{*}{\includegraphics[width=0.9cm]{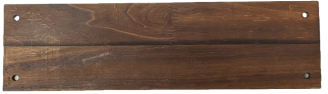}}& \multirow{2}{*}{Plank}
& \textbf{RF} &\textbf{ 98.11} & \textbf{96.5}5 & \textbf{99.17} & \textbf{97.84} \\
\cline{3-7}
&& GB & 97.77 & 95.55 & 99.48 & 97.47 \\
\hline
\multirow{2}{*}{\includegraphics[width=0.65cm]{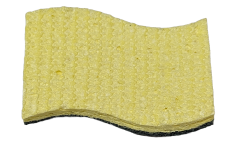}}& \multirow{2}{*}{Sponge}
& \textbf{RF} & \textbf{74.78} & \textbf{62.38} & \textbf{95.64} & \textbf{75.51} \\
\cline{3-7}
&& GB & 43.71 & 41.91 & 99.64 & 59.01 \\
\hline
\multirow{2}{*}{\includegraphics[height=0.65cm]{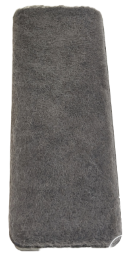}} & \multirow{2}{*}{Duster}
& RF & 99.49 & 99.56 & 99.28 & 99.42 \\
\cline{3-7}
&& \textbf{GB} & \textbf{99.68} & \textbf{99.61} & \textbf{99.66} & \textbf{99.64} \\
\hline
\end{tabularx}
\end{table}

\paragraph{Classifier performance on unseen objects}
Both trained classifier models were tested on previously unencountered objects, selected to evaluate their performance across varied shapes and textures: a soft-covered book, a thermocol duster, a smooth wooden plank, a porous sponge, and a pair of scissors with varying surface features. The results, shown in Table \ref{table:ML_test}, demonstrate the classifiers' effectiveness in accurately detecting slip instances across the five unseen object datasets. The classifiers achieve an accuracy over 99\% for scissors and duster, despite the scissors' abrupt changes in surface area. However, both classifiers encountered difficulty in producing satisfactory results with the sponge. This may be due to the sponge's porous nature, leading to reduced sensor contact, further influenced by its high deformability.

\subsection{Slip Severity Estimation}
\subsubsection{Experiment setup and data collection}
\begin{figure}[b]
    \centering
    \includegraphics[width=0.8\linewidth]{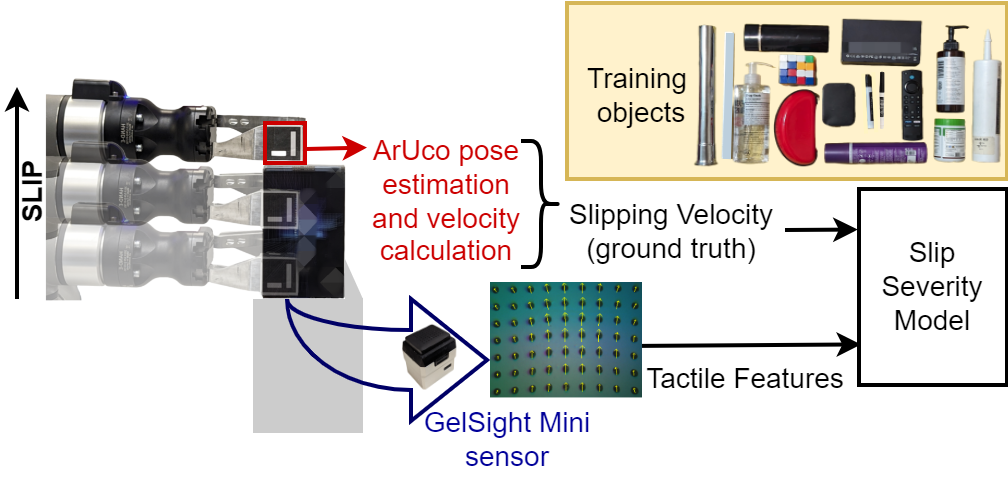}
    \caption{Illustration of the data acquisition framework for the Slip Severity Estimation Model. The gripper, fitted with a GelSight Mini sensor, is programmed to slide over a fixed object. This setup synchronously captures tactile feedback and slip velocity data—the latter serving as ground truth—to train the neural network in assessing slip severity.}
    \label{fig:SS_exp}
\end{figure}

This study aims to assess slip severity in a controlled setting, leveraging a dataset of 15 objects previously selected for slip detection. 
The experiment setup, as shown in Figure \ref{fig:SS_exp} ensured that objects remained stationary, securely fixed to an elevated platform to maintain a constant position relative to the world frame. Emphasis was placed on the gripper's vertical motion along the $y$-axis for specific target velocities. The gripping force was tuned to slightly below the slippage threshold for every object, facilitating the observation of controlled slip events as the gripper moved upward. We focus on linear slipping velocities rather than angular velocities to align with the downstream task of tactile-guided vertical sliding of an object. Slip velocity, denoted as $v_{\text{slip}}$, serves as the ground truth, representing the rate at which an object slips. 


Data collection was carried out for the objects under the following velocity profiles: [0.8, 2.3, 3.8, 4.5, 6.7] \text{cm/s} to observe unidirectional slippage across the sensor surface. To account for variance, we collected data 5 times per velocity for each object. Slip velocities were measured with ArUco markers attached to the gripper and processed with an Exponential Weighted Moving Average (EWMA) filter to reduce noise and smoothen them. The Realsense camera operates at 60Hz, necessitating synchronization of the ArUco pose data with the GelSight sensor. To achieve this, we employ the ROS node \verb|throttle| to publish the pose data into a separate ROS topic at a rate of 25 Hz. Any instance where an object disengaged from the gel surface, ceasing movement, was recorded with a slip velocity of zero to mark the end of slippage.

\subsubsection{Results}
\begin{figure*}[ht!]
\vspace{8pt}
    \centering
    \includegraphics[width=0.75\linewidth]{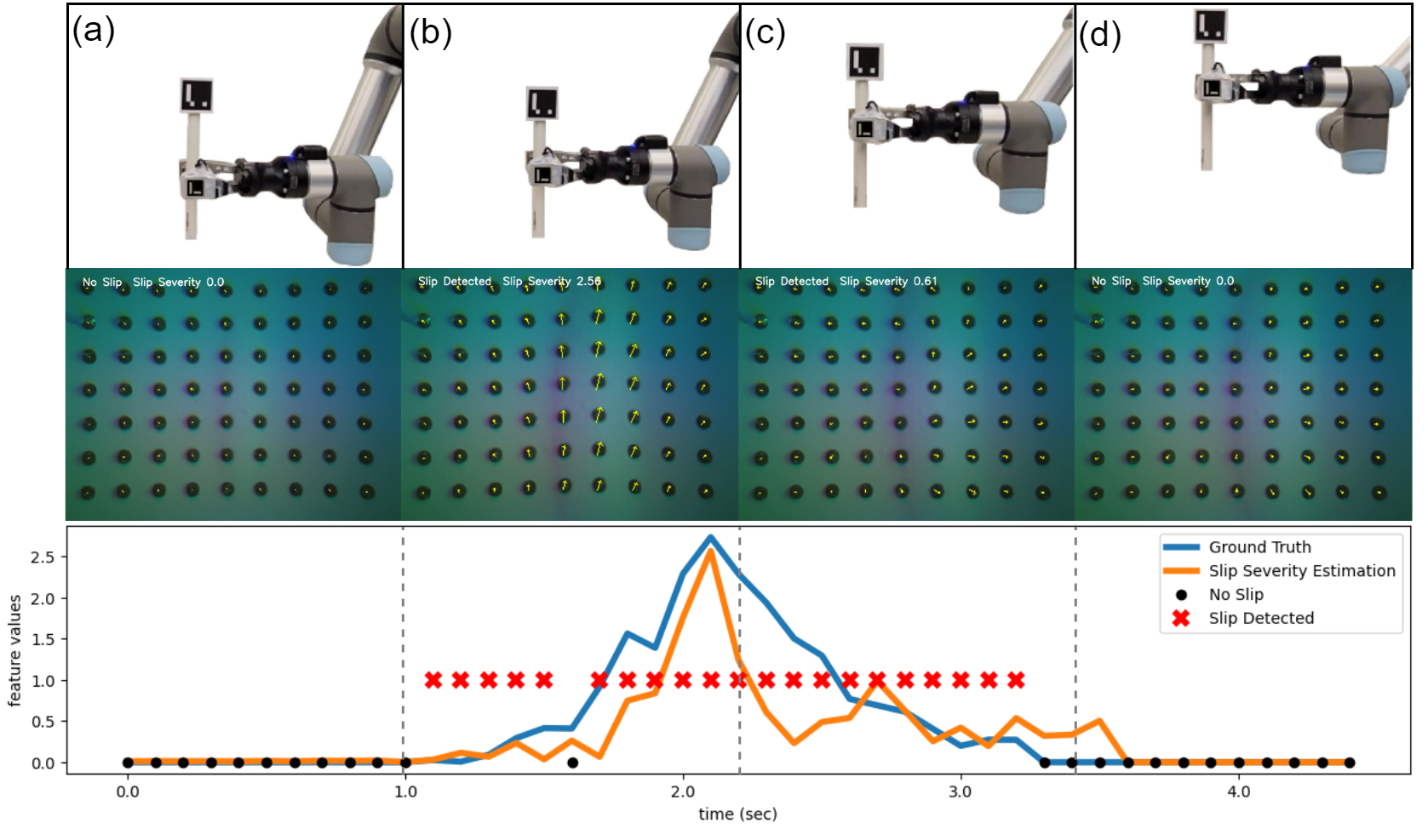}
    \caption{These four snapshots depict a tactile-guided vertical sliding control task. Panel (a) shows the robot grasping the PVC pipe and initiating vertical movement. In panel (b), acceleration leads to increased slip severity, which is concurrently detected and quantified. By panel (c), the robot adjusts its grip to effectively manage slip severity, avoiding overcorrection. Finally, in panel (d), the robot successfully mitigates the slip.}
    \label{fig:pdcontrol}
    \vspace{-10pt}
\end{figure*}
\paragraph{Metrics used for evaluation}
To evaluate the performance of our learned slip severity estimation models, we utilized metrics that capture different aspects of predictive accuracy and error magnitude. 
The chosen metrics are summarized in Table \ref{tab:regression_metrics}.
\begin{table}[thb]
\vspace{5pt}
\centering
\caption{\label{tab:regression_metrics}Estimation metrics summary}
\begin{tabularx}{\columnwidth}{|X|X|X|}
\hline
\textbf{Metric} & \textbf{Definition} & \textbf{Formula} \\ \hline
MAE & Mean absolute diff. & $\frac{1}{n}\sum |y_i - \hat{y}_i|$ \\
RMSE & Square root of MSE & $\sqrt{\frac{1}{n}\sum (y_i - \hat{y_i})^2}$ \\
R\(^2\) & Variance proportion & $1 - \frac{\sum (y_i - \hat{y_i})^2}{\sum (y_i - \bar{y})^2}$ \\
\hline
\end{tabularx}
\vspace{-15pt}
\end{table}
Here, \(y_i\) are the actual values, \(\hat{y}_i\) are the predicted values, \(\bar{y}\) is the mean of actual values, and \(n\) is the number of observations.

\paragraph{Cross-validation for generalization}
The proficiency of LSTM and MLP models in estimating slip severity was assessed using a leave-one-object-out cross-validation approach. Models were trained on data excluding one object and tested against the velocities of the omitted object. The process was iterated for each object, allowing a thorough evaluation. The LSTM model is detailed with 3 layers, 30 hidden units per layer, a 0.2 dropout, and has a sequence length of 5. The MLP model features a three-layer structure, with transitions to 64 and 32 units, supplemented by Layer Normalization and a 0.1 dropout. Metrics discussed above were computed for each test case and then averaged across all objects to gauge the models' overall accuracy, as seen in Table \ref{tab:multiprogram}. Both models demonstrated enhanced precision when utilizing combined feature sets. Specifically, the LSTM model showed marked improvements, with MAE decreasing to 0.26 cm/s, RMSE to 0.77 cm/s, and R\(^2\) rising to 0.75. The MLP model also improved, with MAE reducing to 0.30 cm/s, RMSE to 1.01 cm/s, and R\(^2\) increasing to 0.68. The LSTM model consistently outperformed the MLP, indicating its superior handling of temporal data.

\begin{table}[b]
\vspace{-10pt}
\centering
\caption{Cross-validation results for Slip Severity Estimation}
\setlength\extrarowheight{5pt}
\label{tab:multiprogram}
 \begin{tabularx}{\columnwidth}{| c | c | >{\centering\arraybackslash}X | c | >{\centering\arraybackslash}X |}
  \hline
  \multicolumn{2}{|c|}{{\textbf{Metrics}}} & MAE \textit{(cm/s)} & RMSE \textit{(cm/s)} & R\(^2\)\\
  \hline
  \multirow{3}{*}{LSTM} & Baseline & 0.45 & 1.31 & 0.45\\ \cline{2-5}
  & Proposed & 0.37 & 1.10 & 0.58\\ \cline{2-5}
  & \textbf{Combined} & \textbf{0.26} & \textbf{0.77} & \textbf{0.75}\\
  \hline
  \multirow{3}{*}{MLP} & Baseline & 0.54 & 1.40 & 0.38\\ \cline{2-5}
  & Proposed & 0.43 & 1.21 & 0.51\\ \cline{2-5}
  & \textbf{Combined} & \textbf{0.30} & \textbf{1.01} & \textbf{0.68}\\
  \hline
 \end{tabularx}
\end{table}

\paragraph{Estimation performance on unseen objects}
Table \ref{table:SS_test} details the performance of LSTM and MLP models on the unseen objects with identical slipping velocity profiles, measuring their slip severity estimation accuracy. Performance metrics were calculated for each profile and averaged for an overall effectiveness assessment. The LSTM model consistently surpassed the MLP in MAE and RMSE for all objects, showcasing its enhanced precision and reliability. Notably, the LSTM achieved an impressive R² of 0.80 with the duster, indicating a strong predictive capability. In contrast, the MLP's best performance was an R² of 0.66 with the duster, reflecting moderate efficacy. Particularly, the LSTM showed good performance on a challenging object like the deformable and porous sponge, whereas the MLP struggled significantly on this object.


\subsection{Vertical Sliding Manipulation Control Task}
This experiment integrates learned slip detection and severity models into a unified framework to execute vertical sliding task using a PD (Proportional-Derivative) controller, as detailed in the Methodology section. Due to their performance and generalizability to unseen, challenging objects, we use Random Forest for slip detection and LSTM for slip severity estimation.
\begin{table}[b!]
\vspace{-10pt}
\centering
\setlength\extrarowheight{3pt}
\caption{\label{table:SS_test}Results of trained Slip Severity Estimation models on unseen objects}
\begin{tabularx}{\columnwidth}{|cc|c| >{\centering\arraybackslash}X | >{\centering\arraybackslash}X | >{\centering\arraybackslash}X |}
\hline
\multicolumn{3}{|c|}{{\textbf{Metrics}}} & MAE \textit{(cm/s)} & RMSE \textit{(cm/s)} & R\(^2\)\\
\hline
\multirow{2}{*}{\includegraphics[width = 0.6cm]{pictures_jpeg/book.png}}& 
\multirow{2}{*}{Book}  
& \textbf{LSTM} & \textbf{0.65} & \textbf{0.92} & \textbf{0.65}\\
\cline{3-6}
&& MLP & 0.98 & 1.51 & 0.54\\
\hline
\multirow{2}{*}{\includegraphics[width = 0.7cm]{pictures_jpeg/scissors.png}}& 
\multirow{2}{*}{Scissors}
& \textbf{LSTM} & \textbf{0.53} & \textbf{0.90} & \textbf{0.67}\\
\cline{3-6}
&& MLP & 0.89 & 1.11 & 0.58\\
\hline
\multirow{2}{*}{\centering\includegraphics[width=0.9cm]{pictures_jpeg/woodenpanel.png}}& 
\multirow{2}{*}{Plank} 
&\textbf{LSTM} & \textbf{0.64} & \textbf{1.01} & \textbf{0.68}\\
\cline{3-6}
& &MLP & 1.27 & 1.43 & 0.57\\
\hline
\multirow{2}{*}{\centering\includegraphics[width=0.6cm]{pictures_jpeg/sponge1.png}}& 
\multirow{2}{*}{Sponge} 
&\textbf{LSTM} & \textbf{0.61} & \textbf{1.19} & \textbf{0.60}\\
\cline{3-6}
& &MLP & 1.39 & 1.51 & 0.50\\
\hline
\multirow{2}{*}{\centering\includegraphics[height=0.7cm]{pictures_jpeg/duster2.png}}& 
\multirow{2}{*}{Duster} 
&\textbf{LSTM} & \textbf{0.42} & \textbf{0.68} & \textbf{0.80}\\
\cline{3-6}
& &MLP & 0.67 & 0.93 & 0.66\\
\hline
\end{tabularx}
\vspace{5pt}
\end{table}
The object selected for this demonstration is a smooth PVC pipe, which was not included in the training set of objects. The surface geometry, texture, and inertia of the pipe are unknown to the GelSight sensor. This object was chosen because its slippery surface poses a challenge for slip detection using solely tactile sensors, as seen in Li et al's work \cite{Li2018SlipDW}. The robot is programmed to move at a target velocity of 3.8 cm/s. To calculate ground truth for comparison with slip severity estimation, one ArUco marker is attached to the end effector and another to the top of the object. The ArUco markers' data is published on separate ROS topics at 60Hz, then adjusted to 25Hz through republishing with \verb|throttle|. Additionally, ROS \verb|ApproximateTime| policy is employed to synchronize these topics, facilitating concurrent collection of pose data from both sources. An EWMA filter is applied to smooth the velocities, mirroring the ground truth values used in training the slip severity estimation model. After iterative tuning, the PD controller gains were set to $K_p = 3.10$ and $K_d = 0.42$. Figure \ref{fig:pdcontrol} illustrates the real-time sequence of operations for the experiment. In panel (a), the robot initiates the grasp of the PVC pipe without detecting slip, and the slip severity estimate is 0, as expected. The initial position of the gripper is adjusted to the pipe's diameter plus a tolerance margin, intentionally to induce a slight slip against the GelSight sensor surface. Upon initiating vertical motion, slip is detected, and a real-time slip severity estimate is provided. In panel (b), the slip severity model predicts a value of 2.56 cm/s based on tactile inputs, closely matching the ground truth value of 2.73 cm/s. The tuned PD controller then adjusts the gripper position based on the slip severity error, effectively controlling the slip. Notably, the slip signal is abruptly set to 0 once, marking the sole instance of misclassification during the task. Panel (c) shows that despite ongoing slip detection and slip severity estimation, the estimated value decreases, indicating a reduction in the relative velocity between the object and the sensor. The controller continues to minimize the error until the slip severity estimate reaches 0.00 cm/s, successfully preventing further slip. Even if the slip severity estimator predicts a low value during the transition from (c) to (d), the accuracy of the slip detector prevents overcorrection. The performance on smooth objects like pipes, with an MAE of 0.21 cm/s, RMSE of 0.41 cm/s, and an R² of 0.68, demonstrates strong capability for time-dependent slip severity estimation.

\section{CONCLUSION AND FUTURE WORK}
This study introduces a synergistic approach to detect slips and gauge their severity during the manipulation of objects with diverse shapes and geometries. Utilizing real-time tactile feedback from a GelSight sensor, our system leverages vector field features—velocity, divergence, curl, and normalized contact area, along with their temporal derivatives—to accurately detect slips and predict the velocity of slippage for objects not known \emph{a priori}. Our findings highlight how tactile sensing and feature extraction can enhance the reliability and efficiency of automated object manipulation systems by intelligently mitigating slips.

Our research currently relies on the GelSight Mini sensor for vector flow feature extraction, which, while effective, may present limitations when interfacing with different hardware. We utilize the position-controlled Robotiq Hand-E gripper, which lacks advanced manipulation capabilities. To address this limitation and enhance control precision, we plan to investigate adaptive force control and other advanced algorithms. This direction involves using more capable grippers and training models to estimate both linear and angular slipping velocities from tactile features.
Such advancements would facilitate the development of control strategies that actively manage, rather than merely mitigate, slipping velocity in complex robotic tasks like peg insertion.
Ultimately, these efforts have the potential to significantly improve the dexterity and versatility of robotic systems in handling a variety of objects.



\bibliographystyle{IEEEtran}
\bibliography{cite}
\end{document}